\documentclass[letterpaper, 10 pt, conference]{ieeeconf}  

\IEEEoverridecommandlockouts                              

\usepackage{graphics} 
\usepackage{epsfig} 
\usepackage{amsmath} 
\usepackage{amssymb}  
\usepackage{booktabs} 
\usepackage{hyperref}
\title{\LARGE \bf
Active Vision Might Be All You Need: Exploring Active Vision in Bimanual Robotic Manipulation
}

\author{Ian Chuang*$^{1}$ Andrew Lee*$^{2}$ Dechen Gao$^{2}$  M-Mahdi Naddaf-Sh$^{2}$ Iman Soltani$^\dagger$$^{2}$%
\thanks{*These authors contributed equally to this work.}%
\thanks{$^{1}$University of California, Berkeley $^{2}$University of California, Davis
        }
\thanks{$^{\dagger}$Corresponding email: \href{mailto:isoltani@ucdavis.edu}{isoltani@ucdavis.edu}}
}

\begin{document}

\maketitle
\thispagestyle{empty}
\pagestyle{empty}

\begin{abstract}

Imitation learning has demonstrated significant potential in performing high-precision manipulation tasks using visual feedback. However, it is common practice in imitation learning for cameras to be fixed in place, resulting in issues like occlusion and limited field of view. Furthermore, cameras are often placed in broad, general locations, without an effective viewpoint specific to the robot's task. In this work, we investigate the utility of active vision (AV) for imitation learning and manipulation, in which, in addition to the manipulation policy, the robot learns an AV policy from human demonstrations to dynamically change the robot's camera viewpoint to obtain better information about its environment and the given task. We introduce AV-ALOHA, a new bimanual teleoperation robot system with AV, an extension of the ALOHA 2 robot system, incorporating an additional 7-DoF robot arm that only carries a stereo camera and is solely tasked with finding the best viewpoint. This camera streams stereo video to an operator wearing a virtual reality (VR) headset, allowing the operator to control the camera pose using head and body movements. The system provides an immersive teleoperation experience, with bimanual first-person control, enabling the operator to dynamically explore and search the scene and simultaneously interact with the environment. We conduct imitation learning experiments of our system both in real-world and in simulation, across a variety of tasks that emphasize viewpoint planning. Our results demonstrate the effectiveness of human-guided AV for imitation learning, showing significant improvements over fixed cameras in tasks with limited visibility. Project website: \href{https://soltanilara.github.io/av-aloha/}{https://soltanilara.github.io/av-aloha/}

\end{abstract}

\section{Introduction}

\begin{figure}[t]
    \centering
    \includegraphics[width=\linewidth]{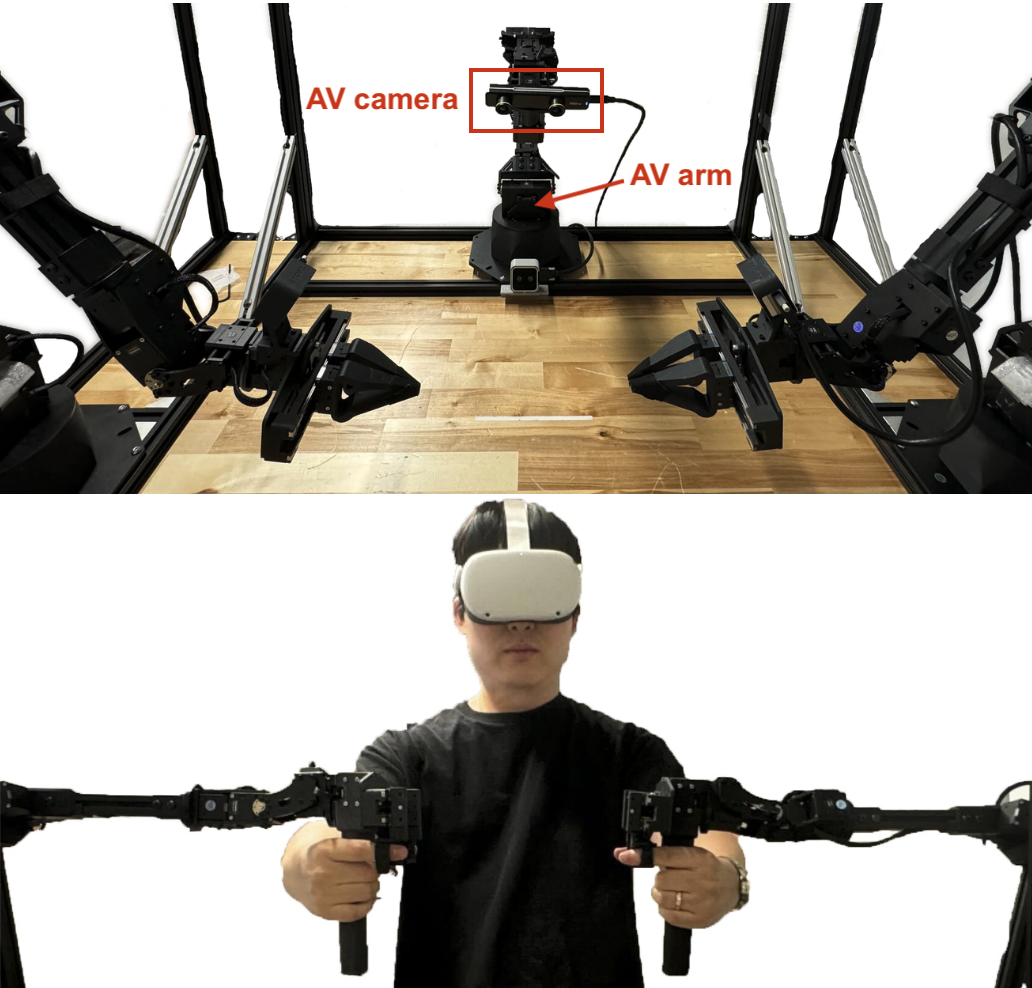}
    \vspace{-20pt}
    \caption{We introduce AV-ALOHA, a bimanual robot system with 7-DoF AV. In this system, a VR headset provides a live feed from the AV camera to the user. The movement of the VR headset controls the AV arm.}
    \vspace{-15pt}
    \label{fig:av-aloha}
\end{figure}

Recent advances in robot learning architectures \cite{zhao_learning_2023, chi2024diffusionpolicy} along with the development of low-cost, open-source methods for easier robot data collection \cite{zhao_learning_2023, chi2024universal}, have led to an accelerated advancement for robot learning using imitation learning methods \cite{lee2024interact, pmlr-v235-lee24y, pmlr-v205-shridhar23a}. End-to-end imitation learning-based approaches offer a scalable and general solution to bimanual tasks that would be very challenging to implement using heuristic-based, task-specific methods. A key feature of these systems is that instead of relying on precise calibration and expensive sensors, these systems can achieve remarkable precision by instead relying on visual feedback from inexpensive cameras \cite{zhao_learning_2023}.

In most robotics implementations, it is common for cameras to be either fixed in place \cite{zhao_learning_2023, collaboration_open_2024,  fu_mobile_2024} or mounted eye-in-hand in combination with a tool like a gripper \cite{chi2024universal, shafiullah2023bringing}. These cameras are typically positioned in a task-agnostic manner, without considering specific visibility requirements of a given task. However, camera placement is crucial for effective learning and execution. With an inadequate camera viewpoint, these models will struggle, especially in situations where the object being handled occludes the camera’s view or when tasks require close-ups of small features \cite{zhao_learning_2023}. 

Consider an assembly task, such as a peg-in-hole scenario where the location of the hole has limited visibility, threading a needle or inserting a key into a lock. These tasks require precision and dexterity, and a camera perspective that captures the relevant features of objects. We believe that a static camera may not always provide an effective viewpoint, whereas a dynamic camera that adjusts its perspective in real-time to the task can offer more flexibility.

For example, consider the bimanual robotic task of inserting a key into a lock. If the lock or key is very small, fixed cameras positioned far from the scene may struggle to provide a clear view, making them difficult to locate and interact with. Additionally, depending on how a robot grasps the lock and key, the view of these objects from the fixed cameras may be occluded by the robot. When attempting to insert the key, the fixed camera may not be positioned at the right angle to see the keyhole. Eye-in-hand cameras attached to the robot's end-effectors can be ineffective since the camera viewpoints are dependent on the robot's task execution. In the example of inserting the key, the eye-in-hand cameras might not have a clear line of sight to the keyhole if the hole is obstructed by the key or the fingers.

To address occlusion and poor perspectives, we propose using AV to control viewpoint to find a more favorable perspective. This is inspired by how we handle manipulation tasks in our daily lives. When inserting a key into a lock, humans adjust their head and gaze to enhance precision. How humans move their gaze independent of their arms, a well studied aspect of human visual feedback \cite{findlay_2003_active, JudgeGrasp, proprioception}, is key to being able to manipulate everyday objects. As we move towards the ultimate goal of human level dexterity for robots, we hypothesize that similar adaptive visual feedback requirements should apply. Many research studies have integrated AV into applications like object tracking or scene reconstruction, demonstrating numerous benefits, including avoiding occlusions, overcoming limited fields of view, and focusing on key points of interest \cite{chen_2011_active, zeng_view_2020}. However, research on AV remains underexplored in robot learning and manipulation.

In our work, we apply and evaluate AV-enhanced imitation learning for dexterous manipulation. Taking advantage of how humans can determine effective viewpoints for completing tasks, we develop a robot system where finding the best camera perspective is directly learned from human demonstrations. We build on the existing ALOHA 2 system \cite{aloha2team2024aloha}, which has two robot arms for bimanual manipulation, and introduce AV-ALOHA, which incorporates an additional 7 DoF arm (AV arm) carrying a stereo camera, dedicated solely for AV. During demonstration, the AV arm is controlled by the user’s head and body movements to dynamically adjust the camera perspective. The user wears a VR headset that streams a live feed from the camera attached to the AV arm, offering an immersive, first-person active sensing experience. During training and data collection, the human operator seamlessly adjusts the camera’s position using the AV arm by naturally moving their body, head, and neck. This allows them to simultaneously execute the task while  attempting to capture an ideal perspective, independent of the manipulator arms. This setup allows for flexible camera movement, mimicking how humans can move their heads to find the best viewpoint. We also developed a simulation environment where users can collect data with just a VR headset and a computer, eliminating the need for physical robot hardware while maintaining the same immersive experience. In addition, in line with the principles of ALOHA, we keep the system open-source and cost-effective, using affordable components and robots. The extension only adds an estimated \$6,600 to the overall cost.

With our teleoperation system, we collect data on a variety of simulation and real-world bimanual manipulation tasks and evaluate a state-of-the-art imitation learning policy, ACT \cite{zhao_learning_2023}, with and without the 7-DoF AV arm. The tasks we chose to test are relatively more challenging compared to those explored in previous publications and may require higher precision as well as be influenced by the selection of camera perspectives. We provide an extensive evaluation of AV in imitation learning and conduct ablation studies highlighting the impact of AV with different camera configurations.

Our contributions are as follows:

\begin{enumerate}

    \item AV-ALOHA, an open-source, low-cost teleoperation system, featuring an additional 7-DoF AV arm, providing a real-time and immersive teleoperation experience.
    \item An open-source simulation environment for AV-ALOHA, featuring new bimanual tasks and publicly available datasets of human demonstrations.
    \item Extensive evaluation of active vision and imitation learning across various simulated and real-world tasks.
    \item Ablation studies highlighting the impact of different camera configurations in combination with AV.
\end{enumerate}

\section{Related Work}

\subsection{Active Vision}

Active vision (AV) was first defined in \cite{aloimonos_active_1988}, where a framework was introduced to more efficiently solve tracking with an active observer. Since then, extensive research has focused on AV, particularly in the domain of object tracking \cite{rivlin_tracking_2000, denzler_3d_tracking_2003, jiang_500-fps_2021, liu_target_2020}. One key area of interest of AV is view planning, which seeks to determine the best sequence of views for a sensor \cite{zeng_view_2020}. Much of this work has been applied to object reconstruction \cite{krainin_autonomous_2011, burusa_attention-driven_2024}, scene reconstruction \cite{davison_simultaneous_2002, davison_real-time_2003, dong_multi_scene_2019}, object recognition \cite{browatzki_2012}, and pose estimation \cite{wu_active_pose_est_2015}.

There are also AV approaches relevant to learning for manipulation. Reinforcement learning policies have been developed for AV, modeling it as a partially observable Markov decision process (POMDP) to handle object manipulation in occluded environments \cite{cheng2018reinforcement, fujita2020distributed}. An energy-based method has been proposed to select the next best view, using a 7-DoF camera attached to an arm to reduce energy and minimize surprise \cite{van_de_maele_active_2021}. Additionally, data-driven AV approaches for grasping focus on selecting perspectives that optimize the grasping policy \cite{natarajan_aiding_2021}. Some works also explore synthetic viewpoint augmentation to scale data for imitation learning, but these methods are not truly AV—they simply aim to increase data rather than find better views \cite{chen_rovi-aug_2024, tian_view-invariant_2024}. Our approach differs in that we focus on human-guided AV. Unlike methods that are either too general to provide task-specific information, or too specialized for tasks like grasping, we propose a scalable approach. By learning from human demonstrations, the teleoperator naturally controls the camera view to find the best perspective.

\subsection{Teleoperation Systems for Data Collection}

Having a robust robot system for collecting human demonstrations is crucial. Recent works and systems have explored innovative approaches to gather robot data. Some systems utilize leader-follower configurations for bimanual control \cite{zhao_learning_2023, fu_mobile_2024, wu_gello_2024}. Others employ VR-based pose estimation or exoskeletons for cartesian space control \cite{lin_learning_2024, yang_ace_2024}. Additionally, some systems focus on simplifying data collection by creating devices that do not require a robot \cite{chi2024universal, shafiullah2023bringing, fang_airexo_2024}. Instead of using parallel jaw grippers, many opt for multifingered hands controlled via hand pose estimation or motion capture gloves \cite{wang_dexcap_2024, qin_anyteleop_2023, iyer_open_2024}.

None of these systems incorporate AV with independent control of camera perspectives. While some, like Open-Television \cite{cheng_open-television_2024}, use immersive first-person teleoperation with a VR headset and an AV two-axis gimbal, they maintain a relatively constrained camera movement, and AV is not their primary focus. Additionally, some industry systems, particularly those with humanoid robots, feature gimbal systems for head movement that only adjust camera direction \cite{noauthor_reachy_nodate, noauthor_sanctuary_nodate}. Unlike these other systems, which have limited ranges of motion and degrees of freedom for AV, our setup uses a dedicated robotic arm, allowing for camera movement in 6-dimensional space much like a human's ability in adapting perspective. This complicates control, but opens new potentials in handling complex robotic tasks.

\subsection{Imitation Learning}

Imitation learning, which involves learning from expert demonstrations, has proven to be an effective approach for robot control. Numerous general architectures have emerged \cite{zhao_learning_2023, chi2024diffusionpolicy, lee2024interact,  pmlr-v235-lee24y} and there are also many efforts to scale up robot data and human demonstrations \cite{collaboration_open_2024, khazatsky2024droidlargescaleinthewildrobot} for generalist language-conditioned policies, \cite{brohan2022rt, zitkovich2023rt, kim2024openvla, octo_2023}. Despite notable advancements, robot learning continues to face significant challenges. Occlusions and the manipulation of small components remain difficult even for state-of-the-art methods \cite{zhao_learning_2023}. The reliance on fixed or eye-in-hand cameras has restricted the ability of robots to effectively handle a range of manipulation tasks. Our work seeks to overcome these limitations by integrating AV into imitation learning to enable robots to tackle new and conventionally difficult tasks. Beyond performance improvements, we aim to deepen our understanding of the challenges posed by high-DoF AV systems and demonstrate its potentials. We further hope to encourage the robotics community to explore these challenges and contribute to the development of next-generation robotic systems with adaptive, human-like vision capabilities.

\section{AV-ALOHA: Description of the Robot System}

\begin{figure*}[t]
    \centering
    \vspace{5pt}
    \includegraphics[scale=.53]{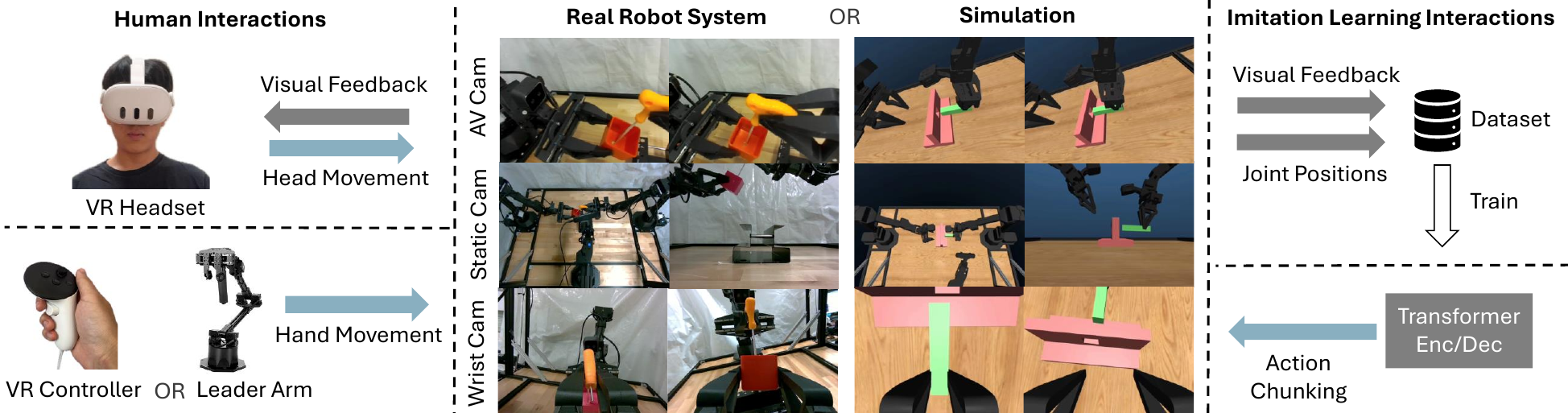}
    \vspace{-5pt}
    \caption{\textbf{Data collection and imitation learning pipeline with AV-ALOHA}: The AV-ALOHA system enables intuitive data collection using a VR headset for AV and either VR controllers or leader arms for manipulation (\textit{left}). This helps capture full body and head movements to teleoperate both our real and simulation system that record video from six different cameras (\textit{middle}) and provide training data for our AV imitation learning policies (\textit{right}).}
    \vspace{-15pt}
    \label{fig:active-vision-system}
\end{figure*}

Our teleoperation, data collection, and autonomous system is illustrated in Figure \ref{fig:active-vision-system}. The system features three Interbotix ViperX-300 6-DoF \cite{noauthor_viperx_nodate} robotic arms: two equipped with grippers for manipulation and an AV arm fitted with a ZED mini stereo camera \cite{noauthor_zed_nodate}, whose movements are controlled by the operator's head movement via a Meta Quest 2 or 3 VR headset \cite{noauthor_meta_nodate}. The two manipulation arms can be controlled either using VR controllers or the original ALOHA leader arms \cite{noauthor_widowx_nodate}, which replicate joint positions. The camera on the AV arm streams two 720p RGB videos to the VR headset.

\subsection{Hardware}

The hardware configuration builds upon the ALOHA 2 \cite{aloha2team2024aloha} setup. We retain the two leader arms and the two follower arms, as well the original four Intel RealSense D405 cameras \cite{noauthor_depth_nodate}. Two cameras are attached eye-in-hand to the follower arms while the other two are fixed to the top and bottom of the setup, providing high- and low-angle perspectives. We introduce a new arm, termed the AV arm, which is an Interbotix ViperX-300 arm equipped with a ZED mini stereo camera on its end effector. We further enhance this arm by converting it from 6-DoF to 7-DoF. The additional degree of freedom addresses the limited range of motion and frequent singularities encountered with the original 6-DoF configuration, significantly expanding the arm’s ability to achieve various camera perspectives. This modification is straightforward, involving only the 3D printing of a small bracket and repurposing the existing gripper motor to provide a mechanism to easily pan the camera.

\subsection{Simulation Environment}

We also developed a simulation environment of the AV-ALOHA system using MuJoCo \cite{todorov2012mujoco}. Building upon the ALOHA 2 model from MuJoCo Menagerie \cite{menagerie2022github}, we incorporated the AV arm, mirroring our real robot system. The data collection process in the simulation uses the same interface as the real robot system, as users can utilize a VR headset to stream stereo video and experience immersive teleoperation within the simulated environment. This simulation was created to offer a systematic and controlled setting for evaluating our AV and imitation learning policies.

\subsection{Teleoperation with VR Headset} 

For the VR headset, we developed a Unity application that interfaces with the robot system via WebRTC \cite{noauthor_aiortcaiortc_2024}. The robot system streams two 720p, 30fps video feeds from the ZED mini’s left and right cameras on the AV arm. These video streams are displayed independently to the operator’s left and right eyes, enhancing immersion and providing a sense of depth and spatial awareness for the teleoperator.

We offer two teleoperation options, both involving the VR headset. The first option uses the VR headset exclusively for control. The headset transmits the tracked poses of the operator's head and hand controllers to the robot system, which then commands the arms accordingly. The grippers of the two arms are operated by pressing the trigger buttons on the hand controllers. The second option integrates the leader arms from the original ALOHA 2 system, allowing for control of the follower arms while the VR headset manages the AV arm. We provide these two interfaces for convenience. For simulation data collection, we chose the VR-only control option for its simplicity and lack of additional hardware requirements. For real-world data collection, we opted for the leader arms for better joint-wise control.

For the VR headset, we obtain the absolute poses of the user's head and hands and convert them to the robot's coordinate frame. The robot arms are initialized at a starting position, and if the first teleoperation option is used, the operator is provided with a visual AR guide for hand placement. Once control begins, all movements are relative to this initial pose. For both teleoperation options, the AV arm receives a target pose from the VR headset. Differential Inverse Kinematics (IK) with Damped Least Squares \cite{buss2004introduction} is used to map this pose to the arm's joint angles. For controlling the manipulation arms with VR hand controllers, we use a Differential IK method with a custom cost function due to the arms frequently approaching joint singularities. This approach evaluates different joint deltas to find those that minimize the cost function. Our cost function penalizes deviations of joints from their center to avoid joint limits and reduces joint displacement to prevent sharp movements.

With our system, attaching the camera to a 7-DoF arm allows for an extended reach and enhanced range of motion. This setup enables us to precisely map the locations of the robot's two end-effectors and the AV arm to correspond exactly with the operator's hand and head positions. This alignment creates a more immersive experience, as the operator and robot directly mirror each other's movements. We believe this provides a more intuitive control and teleoperation experience, making it easier for users to learn and adapt to the system and hence, generate more natural and effective demonstration data for robot learning.

\section{Experiments}

\begin{figure*}[t]
    \centering
    \vspace{5pt}
    \includegraphics[scale=.6]{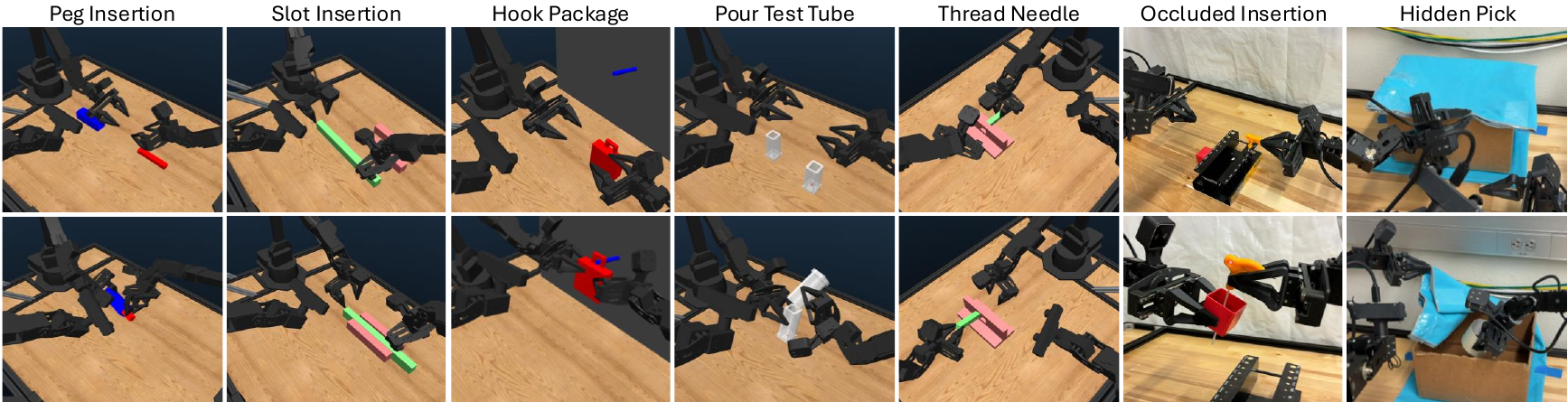}
    \vspace{-8pt}
    \caption{\textbf{Robot Tasks}: We experimented with five simulation and two real-world tasks. Some tasks encourage the robot to adjust its camera perspective.}
    \vspace{-15pt}
    \label{fig:tasks}
\end{figure*}

To evaluate the effectiveness of AV for imitation learning, we adopt a popular imitation learning framework, Action Chunking with Transformers (ACT) \cite{zhao_learning_2023}. We train and evaluate ACT using the library, LeRobot \cite{cadene2024lerobot}, which provides a state of the art implementation of ACT. We use the default implementation of ACT from LeRobot which uses a pretrained ResNet18 \cite{he2016deep} visual backbone. For action chunking, we use a chunk size of 50 for both the simulation and real-world since the real-world data is collected at 50 Hz. Although the robot is teleoperated with cartesian control inputs, we record and train on joint position observations and actions. In contrast to the original ALOHA setup, our action space is expanded to predict 21 joint positions per timestep to accommodate the additional 7-DoF AV arm. For training, we incorporate a learning rate of 2.5e-5 with a batch size of 16 and train for 15625 steps. All other model parameters match the default configuration provided by LeRobot. During training, instead of relying on validation loss, we save multiple checkpoints and directly evaluate the policy on the target environment \cite{mandlekar2021matters}.

We evaluate on five simulation and two real-world tasks. For each task we experiment with different combinations of cameras for the model. AV-ALOHA has six cameras where two are fixed, two are attached eye-in-hand to the wrists of the arm, and two are from the stereo camera attached to the AV arm. We refer to the fixed cameras as \textbf{Static}, eye-in-hand cameras attached to the wrist as \textbf{Wrist}, and AV camera as \textbf{AV}. We experiment on all 7 different possible combinations of these cameras and evaluate the success rate. 

For each task, we collected 50 episodes of human demonstrations with all three arms while recording from all cameras. Then we selectively train with the specific camera configuration. We decided not to collect separate data for each different camera configuration to keep the trajectories of the data consistent for training between different configurations. However, for configurations that don't require an AV camera, we acknowledge that the AV arm could potentially show up in the frame of the other cameras while also requiring additional control inputs for the policy. Thus, in the real world experiment, even when the policy doesn't use the AV camera we still have it control the AV arm to keep the data consistent. However for simulation experiments, we can render and record the same trajectories twice to both include and not include the AV arm. Therefore, in simulation, we can train and evaluate with and without the AV arm. 

\subsection{Tasks}

All seven tasks, depicted in Figure \ref{fig:tasks}, are designed to conduct bimanual manipulation. Three of these tasks (Group 1)—\textbf{Peg Insertion}, \textbf{Slot Insertion}, and \textbf{Hook Package}—can be completed without the need for AV, as the standard ALOHA camera setup (including static and wrist cameras) is sufficient for task execution. In contrast, the remaining four tasks (Group 2)—\textbf{Pour Test Tube}, \textbf{Thread Needle}, \textbf{Occluded Insertion}, and \textbf{Hidden Pick}—are designed to potentially benefit from improved camera perspectives provided by AV. By evaluating both scenarios, we gain insight into the advantages of AV in the latter group, while, through the former, we also identify any potential drawbacks, such as increased complexity from operating in a larger action space, or processing additional cameras inputs.

In simulation, initial object positions are randomized in a small rectangular region. For real-world tasks, we manually introduce variations in initial object positions. For all tasks, the three arms always start from a fixed initial position.

\textbf{Peg Insertion} is a simulation task adapted from the original ALOHA paper \cite{zhao_learning_2023} to our new simulation where the right arm needs to grasp a peg/stick and the left arm needs to grab a socket. The two arms then coordinate to insert the peg into the socket. \textbf{Slot Insertion} is a simulation task adapted from \cite{lee2024interact} where the two arms need to grasp a long stick from both ends and insert it into a slot. \textbf{Hook Package} is a new simulation task in which both arms work together to grasp a package or box. The package has a tab with a hole, and the goal is to hang the package on a hook attached to a wall. We categorize \textbf{Peg Insertion} and \textbf{Slot Insertion} into Group 1, as these tasks have previously been demonstrated autonomously using a single static camera, as shown in \cite{zhao_learning_2023, lee2024interact}. \textbf{Hook Package} is also placed in Group 1, as the package and hook remain visible to the static cameras.

\textbf{Pour Test Tube} is a simulation task involving two slim tubes, where one tube contains a small marble. The two arms need to grasp the tubes and the right arm pours the marble from one tube into the other tube. We believe this task may require AV to look closely at the ball while pouring for more precision, similar to how a human might accomplish this task. \textbf{Thread Needle} is a new simulation task where the right arm grasps a needle and threads it through a hole on an object for the left arm to grab and pull it out on the opposite side of the hole. The object with the hole is placed so that its hole has limited visibility from the static cameras in the setup. \textbf{Occluded Insertion} is a real-world task where the right arm grabs a long allen key from a tray and the left arm grabs a small container with a hole at the bottom of the container. The right arm then needs to insert the end of the allen key into the hole of the container. However, since the hole is located at the bottom of the container, it may be occluded by the sides of the container, making the hole difficult to see without a proper camera perspective adjustment. \textbf{Hidden Pick} is a real-world task involving an object inside a box hidden by a towel that covers the top of the box. The left arm should grasp and move the towel, revealing the object. The right arm then reaches into the box, picks up the object, and places it outside. These four tasks are categorized into Group 2 because they involve occlusions or require a level of precision that could be addressed with AV.

For the \textit{hidden pick} task, data collection and experiments were carried out with an updated version of AV-ALOHA equipped with a different stereo camera with a wider field of view. Data for this task was recorded at 33fps and the trained ACT policy used a chunk size of 33.
 
\section{Results and Discussion}

\begin{table*}[t!]
\vspace{5pt}
\caption{Success rates (\%) of ACT policy across different tasks and cameras. Fixed cameras are denoted as \textbf{Static}, eye-in-hand cameras as \textbf{Wrist}, and AV camera as \textbf{AV}. Each task consists of two stages indicating partial and full success. Group 1 includes tasks that may not need AV, while Group 2 involves occlusions or high precision that could benefit from AV.}
\vspace{-5pt}
\centering
\resizebox{\textwidth}{!}{
\begin{tabular}{c|cccccc|cccccccc}
\toprule
\multicolumn{1}{c}{} &
\multicolumn{6}{c}{\textbf{Group 1}} & \multicolumn{8}{c}{\textbf{Group 2}} \\
\cmidrule(lr){2-7} \cmidrule(lr){8-15}
 & \multicolumn{2}{c}{\textbf{Peg Insertion}} 
 & \multicolumn{2}{c}{\textbf{Slot Insertion}} 
 & \multicolumn{2}{c}{\textbf{Hook Package}} 
 & \multicolumn{2}{c}{\textbf{Pour Test Tube}} 
 & \multicolumn{2}{c}{\textbf{Thread Needle}} 
 & \multicolumn{2}{c}{\textbf{Occluded Insertion}} 
 & \multicolumn{2}{c}{\textbf{Hidden Pick}} \\ 
\cmidrule(lr){2-3}
\cmidrule(lr){4-5}
\cmidrule(lr){6-7}
\cmidrule(lr){8-9}
\cmidrule(lr){10-11}
\cmidrule(lr){12-13}
\cmidrule(lr){14-15}
\textbf{Camera Config} & Grasp & Insert & Grasp & Insert & Grasp & Hook & Grasp & Pour & Grasp & Thread & Grasp & Insert & Reveal & Pick \\
\midrule
\textbf{AV} & 74 & 42 & 88 & 50 & 100 & 22 & 66 & \textbf{14} & 98 & \textbf{52} & 60 & 20 & 95 & 55 \\
\textbf{AV} + Static & 84 & 46 & 100 & 62 & 100 & 34 & 50 & 10 & 98 & 26 & 20 & 0 & 90 & 45 \\
\textbf{AV} + Wrist & 82 & 34 & 96 & 44 & 100 & 22 & 70 & \textbf{14} & 92 & \textbf{52} & 95 & \textbf{30} & 95 & \textbf{60} \\
\textbf{AV} + Static + Wrist & 78 & 36 & 100 & 36 & 100 & 24 & 36 & 8 & 90 & 40 & 40 & 5 & 55 & 15 \\
Static & 84 & \textbf{48} & 98 & 66 & 100 & \textbf{44} & 44 & 8 & 88 & 30 & 85 & 20 & 100 & 40 \\
Static + Wrist & 88 & 40 & 100 & \textbf{78} & 100 & 30 & 46 & 6 & 38 & 22 & 100 & 15 & 100 & 35 \\
Wrist & 84 & 42 & 98 & 44 & 92 & 8 & 44 & 10 & 94 & 44 & 60 & 15 & 95 & 35 \\
\bottomrule
\end{tabular}
}
\label{table:results}
\vspace{-15pt}
\end{table*}

For simulation tasks, we evaluate each camera configuration using 12 different policy checkpoints from training, rolling out each checkpoint 50 times, and report the results for the best-performing checkpoint. For real-world tasks, we roll out the final checkpoint 20 times and report the success rate for each configuration. Results are shown in Table \ref{table:results}.

For group 1 tasks, non-AV setups achieved higher success rate on two tasks, (\textit{slot insertion} and \textit{hook package}), and for the \textit{peg insertion} task, results were comparable between the two setups. This indicates that for those cases where AV is not necessarily advantageous, the added complexity from variation in viewpoint can negatively impact performance. This is further supported even in non-AV scenarios that use wrist cameras, which also introduce a dynamic viewpoint. For example for \textit{hook package}, while the highest performance is obtained when using static cameras alone, inclusion of the wrist cameras hinders the success rate.    

For group 2 tasks, we found that setups with AV showed improvements over non-AV setups. For \textit{thread needle} and \textit{pour test tube}, camera configurations with AV performed very well, with the AV and AV + wrist configurations achieving the two highest success rates on these tasks. For \textit{thread needle}, the hole is not easily visible from the static cameras and the AV camera is able to get a good perspective of the hole, which is crucial for inserting the needle through the hole. When comparing the perspectives of the wrist and AV cameras, although wrist cameras are able to provide a reasonable view of each side of the hole, the AV camera can provide a more holistic view of all the components involved. For \textit{pour test tube} due to the slim design of the tubes and small marble size, the AV camera can focus and zoom in on the openings of the tubes and provide a clearer view of the marble for better precision in the task.

For our real-world tasks, the AV + wrist combination performed exceptionally well compared to other configurations. For \textit{hidden pick}, we noticed the box and towel caused major occlusions for the static and wrist camera views, whereas AV was able to find a clearer view showing the entire inside of the box. In \textit{occluded insertion}, during inference with AV + wrist, we observed that the arm with the peg appeared to use visual feedback to align the peg and the hole, whereas in configurations without AV, the arm with the peg would forcefully press the peg against the container. From this behavior, we infer that AV enhances precision in tasks where visual feedback of small and intricate details is crucial.

We observed that AV alone achieved the top success rate on \textit{pour test tube} and \textit{thread needle} and performed well for the other tasks. This suggests AV might be sufficient enough for adequate performance across multiple tasks, without the need to install multiple fixed cameras around the scene. 

Results indicate that using static cameras only performed best on two of the group 1 tasks, i.e. \textit{peg insertion}, and \textit{hook package}. Investigation of the camera feeds indicated that in these tasks, static camera perspective provided all the necessary visual information to complete the task. Additionally, since these cameras do not move, they provide a more stable and predictable vision input, whereas AV and wrist camera images change significantly following the control inputs, resulting in a more complex visual feedback/control system. We also hypothesize that fixed cameras benefit from a ``fixed coordinate system'' where objects in a particular location are always located in the same corresponding pixel positions, making it easier for the model to interpret their locations. In contrast, the moving cameras, would introduce additional complexity in tracking and interpreting object locations on the scene. In such cases, where static cameras suffice to execute the task, we noticed that adding more moving cameras can deteriorate the results. 

We further observed that in 5 out of 7 tasks, AV + static combination, outperforms static + wrist. These two configurations are similar in terms of data complexity and network architecture, but in AV + static scenarios the perspective control is decoupled from the object manipulation, unlike static + wrist. It is noted that AV + static setup further complicates the control requirements and hence, may gain additional benefit from more complex control architectures.     


Another interesting result was that using all the cameras simultaneously did not perform well across the tasks and never ranked in the top three for any task. One explanation from the observations is that adding more cameras can actually hurt performance if the additional cameras do not provide significant new information. This result can be further attributed to the significantly larger action space and more complex, decoupled nature of vision control in AV, which may necessitate more complex control architectures, more training data, or additional training. 


\section{Conclusion}

In this work, we introduced a novel robotic setup featuring a 7-DoF AV arm. Through extensive experiments, we demonstrated that AV can significantly improve imitation learning, particularly in tasks that can benefit from proper selection of the camera perspective. Our results suggest that AV may be sufficient to provide the necessary visual feedback for successful task execution across a broad range of tasks, thereby improving the generalizability of robotic platforms and potentially reducing the need for additional, task-specific camera setups. This approach is inspired by how humans dynamically adjust their perspectives using head, neck, and waist movements to optimize their view during manipulation tasks.

To explore the utility of AV, we conducted an in-depth analysis of its role alongside static and conventional moving (eye-in-hand) camera setups. While AV shows significant potential, it also introduces complexities that warrant further research, particularly in the development of control architectures capable of managing the decoupled nature of visual feedback and control, the expanded action space, and the method's susceptibility to distribution shift. Additionally, our results raise important questions regarding the data and training requirements for such systems, highlighting the need for continued investigation into how AV can be integrated into robotic platforms. 


Moving forward, we aim to contribute to the development of more generalizable, human-like robotic platforms where task-specific visual data is delivered in a targeted, controlled manner, with redundant information filtered at the sensing level rather than at the computational level. The results presented in this paper, along with open-sourcing AV-ALOHA hardware and software, represent important first steps toward achieving this goal. We hope that this work inspires further research in this direction, ultimately leading to more efficient and adaptable robotic systems.


\addtolength{\textheight}{0cm}   





\section*{Acknowledgement}

We thank Mohnish Gopi for assisting with the VR app and Soumyajit Ganguly for helping on the design tasks.


\bibliography{main}

\begin{thebibliography}{10}
\providecommand{\url}[1]{#1}
\csname url@rmstyle\endcsname
\providecommand{\newblock}{\relax}
\providecommand{\bibinfo}[2]{#2}
\providecommand\BIBentrySTDinterwordspacing{\spaceskip=0pt\relax}
\providecommand\BIBentryALTinterwordstretchfactor{4}
\providecommand\BIBentryALTinterwordspacing{\spaceskip=\fontdimen2\font plus
\BIBentryALTinterwordstretchfactor\fontdimen3\font minus \fontdimen4\font\relax}
\providecommand\BIBforeignlanguage[2]{{%
\expandafter\ifx\csname l@#1\endcsname\relax
\typeout{** WARNING: IEEEtran.bst: No hyphenation pattern has been}%
\typeout{** loaded for the language `#1'. Using the pattern for}%
\typeout{** the default language instead.}%
\else
\language=\csname l@#1\endcsname
\fi
#2}}

\bibitem{zhao_learning_2023}
T.~Z. Zhao, V.~Kumar, S.~Levine, and C.~Finn, ``Learning fine-grained bimanual manipulation with low-cost hardware,'' \emph{RSS}, 2023.

\bibitem{chi2024diffusionpolicy}
C.~Chi, Z.~Xu, S.~Feng, E.~Cousineau, Y.~Du, B.~Burchfiel, R.~Tedrake, and S.~Song, ``Diffusion policy: Visuomotor policy learning via action diffusion,'' \emph{The International Journal of Robotics Research}, 2024.

\bibitem{chi2024universal}
C.~Chi, Z.~Xu, C.~Pan, E.~Cousineau, B.~Burchfiel, S.~Feng, R.~Tedrake, and S.~Song, ``Universal manipulation interface: Inthe-wild robot teaching without in-the-wild robots,'' \emph{arXiv preprint arXiv:2402.10329}, 2024.

\bibitem{lee2024interact}
A.~Lee, I.~Chuang, L.-Y. Chen, and I.~Soltani, ``Interact: Inter-dependency aware action chunking with hierarchical attention transformers for bimanual manipulation,'' \emph{arXiv preprint arXiv:2409.07914}, 2024.

\bibitem{pmlr-v235-lee24y}
S.~Lee, Y.~Wang, H.~Etukuru, H.~J. Kim, N.~M.~M. Shafiullah, and L.~Pinto, ``Behavior generation with latent actions,'' \emph{arXiv preprint arXiv:2403.03181}, 2024.

\bibitem{pmlr-v205-shridhar23a}
M.~Shridhar, L.~Manuelli, and D.~Fox, ``Perceiver-actor: A multi-task transformer for robotic manipulation,'' in \emph{Conference on Robot Learning}.\hskip 1em plus 0.5em minus 0.4em\relax PMLR, 2023, pp. 785--799.

\bibitem{collaboration_open_2024}
O.-X.~E. Collaboration, A.~Padalkar, A.~Pooley, A.~Jain, A.~Bewley, A.~Herzog, A.~Irpan, A.~Khazatsky, A.~Rai, A.~Singh, \emph{et~al.}, ``Open x-embodiment: Robotic learning datasets and rt-x models,'' \emph{arXiv preprint arXiv:2310.08864}, 2023.

\bibitem{fu_mobile_2024}
Z.~Fu, T.~Z. Zhao, and C.~Finn, ``Mobile aloha: Learning bimanual mobile manipulation with low-cost whole-body teleoperation,'' \emph{arXiv preprint arXiv:2401.02117}, 2024.

\bibitem{shafiullah2023bringing}
N.~M.~M. Shafiullah, A.~Rai, H.~Etukuru, Y.~Liu, I.~Misra, S.~Chintala, and L.~Pinto, ``On bringing robots home,'' \emph{arXiv preprint arXiv:2311.16098}, 2023.

\bibitem{findlay_2003_active}
J.~M. Findlay and I.~D. Gilchrist, \emph{Active vision: The psychology of looking and seeing}.\hskip 1em plus 0.5em minus 0.4em\relax Oxford University Press, 2003, no.~37.

\bibitem{JudgeGrasp}
G.~Maiello, M.~Schepko, L.~K. Klein, V.~C. Paulun, and R.~W. Fleming, ``Humans can visually judge grasp quality and refine their judgments through visual and haptic feedback,'' \emph{Frontiers in Neuroscience}, vol.~14, p. 591898, 2021.

\bibitem{proprioception}
R.~Goodman and L.~Tremblay, ``Using proprioception to control ongoing actions: dominance of vision or altered proprioceptive weighing?'' \emph{Experimental Brain Research}, vol. 236, p. 1897–1910, 04 2018.

\bibitem{chen_2011_active}
S.~Chen, Y.~Li, and N.~M. Kwok, ``Active vision in robotic systems: A survey of recent developments,'' \emph{The International Journal of Robotics Research}, vol.~30, no.~11, pp. 1343--1377, 2011.

\bibitem{zeng_view_2020}
R.~Zeng, Y.~Wen, W.~Zhao, and Y.-J. Liu, ``View planning in robot active vision: A survey of systems, algorithms, and applications,'' \emph{Computational Visual Media}, vol.~6, pp. 225--245, 2020.

\bibitem{aloha2team2024aloha}
\BIBentryALTinterwordspacing
{ALOHA 2 Team}, ``Aloha 2: An enhanced low-cost hardware for bimanual teleoperation,'' 2024. [Online]. Available: \url{https://aloha-2.github.io/}
\BIBentrySTDinterwordspacing

\bibitem{aloimonos_active_1988}
J.~Aloimonos, I.~Weiss, and A.~Bandyopadhyay, ``Active vision,'' \emph{International journal of computer vision}, vol.~1, pp. 333--356, 1988.

\bibitem{rivlin_tracking_2000}
E.~Rivlin and H.~Rotstein, ``Control of a camera for active vision: Foveal vision, smooth tracking and saccade,'' \emph{International Journal of Computer Vision}, vol.~39, pp. 81--96, 2000.

\bibitem{denzler_3d_tracking_2003}
Denzler, Zobel, and Niemann, ``Information theoretic focal length selection for real-time active 3d object tracking,'' in \emph{Proceedings Ninth IEEE International Conference on Computer Vision}.\hskip 1em plus 0.5em minus 0.4em\relax IEEE, 2003, pp. 400--407.

\bibitem{jiang_500-fps_2021}
M.~Jiang, R.~Sogabe, K.~Shimasaki, S.~Hu, T.~Senoo, and I.~Ishii, ``500-fps omnidirectional visual tracking using three-axis active vision system,'' \emph{IEEE Transactions on Instrumentation and Measurement}, vol.~70, pp. 1--11, 2021.

\bibitem{liu_target_2020}
Y.~Liu, P.~Sun, and A.~Namiki, ``Target tracking of moving and rotating object by high-speed monocular active vision,'' \emph{IEEE Sensors Journal}, vol.~20, no.~12, pp. 6727--6744, 2020.

\bibitem{krainin_autonomous_2011}
M.~Krainin, B.~Curless, and D.~Fox, ``Autonomous generation of complete 3d object models using next best view manipulation planning,'' in \emph{2011 IEEE international conference on robotics and automation}.\hskip 1em plus 0.5em minus 0.4em\relax IEEE, 2011, pp. 5031--5037.

\bibitem{burusa_attention-driven_2024}
A.~K. Burusa, E.~J. van Henten, and G.~Kootstra, ``Attention-driven next-best-view planning for efficient reconstruction of plants and targeted plant parts,'' \emph{Biosystems Engineering}, vol. 246, pp. 248--262, 2024.

\bibitem{davison_simultaneous_2002}
A.~J. Davison and D.~W. Murray, ``Simultaneous localization and map-building using active vision,'' \emph{IEEE transactions on pattern analysis and machine intelligence}, vol.~24, no.~7, pp. 865--880, 2002.

\bibitem{davison_real-time_2003}
A.~J. Davison, W.~W. Mayol, and D.~W. Murray, ``Real-time localization and mapping with wearable active vision,'' in \emph{The Second IEEE and ACM International Symposium on Mixed and Augmented Reality, 2003. Proceedings.}\hskip 1em plus 0.5em minus 0.4em\relax IEEE, 2003, pp. 18--27.

\bibitem{dong_multi_scene_2019}
S.~Dong, K.~Xu, Q.~Zhou, A.~Tagliasacchi, S.-Q. Xin, M.~Nießner, and B.~Chen, ``Multi-robot collaborative dense scene reconstruction,'' \emph{ACM Transactions on Graphics}, vol.~38, pp. 1--16, 07 2019.

\bibitem{browatzki_2012}
B.~Browatzki, V.~Tikhanoff, G.~Metta, H.~H. B{\"u}lthoff, and C.~Wallraven, ``Active object recognition on a humanoid robot,'' in \emph{2012 IEEE international conference on robotics and automation}.\hskip 1em plus 0.5em minus 0.4em\relax IEEE, 2012, pp. 2021--2028.

\bibitem{wu_active_pose_est_2015}
K.~Wu, R.~Ranasinghe, and G.~Dissanayake, ``Active recognition and pose estimation of household objects in clutter,'' in \emph{2015 IEEE International Conference on Robotics and Automation (ICRA)}.\hskip 1em plus 0.5em minus 0.4em\relax IEEE, 2015, pp. 4230--4237.

\bibitem{cheng2018reinforcement}
R.~Cheng, A.~Agarwal, and K.~Fragkiadaki, ``Reinforcement learning of active vision for manipulating objects under occlusions,'' in \emph{Conference on Robot Learning}.\hskip 1em plus 0.5em minus 0.4em\relax PMLR, 2018, pp. 422--431.

\bibitem{fujita2020distributed}
Y.~Fujita, K.~Uenishi, A.~Ummadisingu, P.~Nagarajan, S.~Masuda, and M.~Y. Castro, ``Distributed reinforcement learning of targeted grasping with active vision for mobile manipulators,'' in \emph{2020 IEEE/RSJ International Conference on Intelligent Robots and Systems (IROS)}.\hskip 1em plus 0.5em minus 0.4em\relax IEEE, 2020, pp. 9712--9719.

\bibitem{van_de_maele_active_2021}
T.~Van~de Maele, T.~Verbelen, O.~{\c{C}}atal, C.~De~Boom, and B.~Dhoedt, ``Active vision for robot manipulators using the free energy principle,'' \emph{Frontiers in neurorobotics}, vol.~15, p. 642780, 2021.

\bibitem{natarajan_aiding_2021}
S.~Natarajan, G.~Brown, and B.~Calli, ``Aiding grasp synthesis for novel objects using heuristic-based and data-driven active vision methods,'' \emph{Frontiers in Robotics and AI}, vol.~8, p. 696587, 2021.

\bibitem{chen_rovi-aug_2024}
L.~Y. Chen, C.~Xu, K.~Dharmarajan, Z.~Irshad, R.~Cheng, K.~Keutzer, M.~Tomizuka, Q.~Vuong, and K.~Goldberg, ``Rovi-aug: Robot and viewpoint augmentation for cross-embodiment robot learning,'' \emph{arXiv preprint arXiv:2409.03403}, 2024.

\bibitem{tian_view-invariant_2024}
S.~Tian, B.~Wulfe, K.~Sargent, K.~Liu, S.~Zakharov, V.~Guizilini, and J.~Wu, ``View-invariant policy learning via zero-shot novel view synthesis,'' \emph{arXiv preprint arXiv:2409.03685}, 2024.

\bibitem{wu_gello_2024}
P.~Wu, Y.~Shentu, Z.~Yi, X.~Lin, and P.~Abbeel, ``Gello: A general, low-cost, and intuitive teleoperation framework for robot manipulators,'' \emph{arXiv preprint arXiv:2309.13037}, 2023.

\bibitem{lin_learning_2024}
T.~Lin, Y.~Zhang, Q.~Li, H.~Qi, B.~Yi, S.~Levine, and J.~Malik, ``Learning visuotactile skills with two multifingered hands,'' \emph{arXiv preprint arXiv:2404.16823}, 2024.

\bibitem{yang_ace_2024}
S.~Yang, M.~Liu, Y.~Qin, R.~Ding, J.~Li, X.~Cheng, R.~Yang, S.~Yi, and X.~Wang, ``Ace: A cross-platform visual-exoskeletons system for low-cost dexterous teleoperation,'' \emph{arXiv preprint arXiv:2408.11805}, 2024.

\bibitem{fang_airexo_2024}
H.~Fang, H.-S. Fang, Y.~Wang, J.~Ren, J.~Chen, R.~Zhang, W.~Wang, and C.~Lu, ``Airexo: Low-cost exoskeletons for learning whole-arm manipulation in the wild,'' in \emph{2024 IEEE International Conference on Robotics and Automation (ICRA)}.\hskip 1em plus 0.5em minus 0.4em\relax IEEE, 2024, pp. 15\,031--15\,038.

\bibitem{wang_dexcap_2024}
C.~Wang, H.~Shi, W.~Wang, R.~Zhang, L.~Fei-Fei, and C.~K. Liu, ``Dexcap: Scalable and portable mocap data collection system for dexterous manipulation,'' \emph{arXiv preprint arXiv:2403.07788}, 2024.

\bibitem{qin_anyteleop_2023}
Y.~Qin, W.~Yang, B.~Huang, K.~Van~Wyk, H.~Su, X.~Wang, Y.-W. Chao, and D.~Fox, ``Anyteleop: A general vision-based dexterous robot arm-hand teleoperation system,'' \emph{arXiv preprint arXiv:2307.04577}, 2023.

\bibitem{iyer_open_2024}
A.~Iyer, Z.~Peng, Y.~Dai, I.~Guzey, S.~Haldar, S.~Chintala, and L.~Pinto, ``Open teach: A versatile teleoperation system for robotic manipulation,'' \emph{arXiv preprint arXiv:2403.07870}, 2024.

\bibitem{cheng_open-television_2024}
X.~Cheng, J.~Li, S.~Yang, G.~Yang, and X.~Wang, ``Open-television: teleoperation with immersive active visual feedback,'' \emph{arXiv preprint arXiv:2407.01512}, 2024.

\bibitem{noauthor_reachy_nodate}
\BIBentryALTinterwordspacing
``\BIBforeignlanguage{en-us}{Reachy by {Pollen} {Robotics}, an open source programmable humanoid robot}.'' [Online]. Available: \url{https://www.pollen-robotics.com/}
\BIBentrySTDinterwordspacing

\bibitem{noauthor_sanctuary_nodate}
\BIBentryALTinterwordspacing
``\BIBforeignlanguage{en-US}{Sanctuary ai}.'' [Online]. Available: \url{https://sanctuary.ai/}
\BIBentrySTDinterwordspacing

\bibitem{khazatsky2024droidlargescaleinthewildrobot}
A.~Khazatsky, K.~Pertsch, S.~Nair, A.~Balakrishna, S.~Dasari, S.~Karamcheti, S.~Nasiriany, M.~K. Srirama, L.~Y. Chen, K.~Ellis, \emph{et~al.}, ``Droid: A large-scale in-the-wild robot manipulation dataset,'' \emph{arXiv preprint arXiv:2403.12945}, 2024.

\bibitem{brohan2022rt}
A.~Brohan, N.~Brown, J.~Carbajal, Y.~Chebotar, J.~Dabis, C.~Finn, K.~Gopalakrishnan, K.~Hausman, A.~Herzog, J.~Hsu, \emph{et~al.}, ``Rt-1: Robotics transformer for real-world control at scale,'' \emph{arXiv preprint arXiv:2212.06817}, 2022.

\bibitem{zitkovich2023rt}
B.~Zitkovich, T.~Yu, S.~Xu, P.~Xu, T.~Xiao, F.~Xia, J.~Wu, P.~Wohlhart, S.~Welker, A.~Wahid, \emph{et~al.}, ``Rt-2: Vision-language-action models transfer web knowledge to robotic control,'' in \emph{Conference on Robot Learning}.\hskip 1em plus 0.5em minus 0.4em\relax PMLR, 2023, pp. 2165--2183.

\bibitem{kim2024openvla}
M.~J. Kim, K.~Pertsch, S.~Karamcheti, T.~Xiao, A.~Balakrishna, S.~Nair, R.~Rafailov, E.~Foster, G.~Lam, P.~Sanketi, \emph{et~al.}, ``Openvla: An open-source vision-language-action model,'' \emph{arXiv preprint arXiv:2406.09246}, 2024.

\bibitem{octo_2023}
O.~M. Team, D.~Ghosh, H.~Walke, K.~Pertsch, K.~Black, O.~Mees, S.~Dasari, J.~Hejna, T.~Kreiman, C.~Xu, \emph{et~al.}, ``Octo: An open-source generalist robot policy,'' \emph{arXiv preprint arXiv:2405.12213}, 2024.

\bibitem{noauthor_viperx_nodate}
\BIBentryALTinterwordspacing
``\BIBforeignlanguage{en}{{ViperX} 300 {S}}.'' [Online]. Available: \url{https://www.trossenrobotics.com/viperx-300}
\BIBentrySTDinterwordspacing

\bibitem{noauthor_zed_nodate}
\BIBentryALTinterwordspacing
``\BIBforeignlanguage{en}{{ZED} {Mini} {Stereo} {Camera} {\textbar} {Stereolabs}}.'' [Online]. Available: \url{https://www.stereolabs.com/store/products/zed-mini}
\BIBentrySTDinterwordspacing

\bibitem{noauthor_meta_nodate}
\BIBentryALTinterwordspacing
``\BIBforeignlanguage{en}{Meta {Quest} {VR} {Headsets}, {Accessories} \& {Equipment} {\textbar} {Meta} {Quest}}.'' [Online]. Available: \url{https://www.meta.com/quest/}
\BIBentrySTDinterwordspacing

\bibitem{noauthor_widowx_nodate}
\BIBentryALTinterwordspacing
``\BIBforeignlanguage{en}{{WidowX} 250 {S}}.'' [Online]. Available: \url{https://www.trossenrobotics.com/widowx-250}
\BIBentrySTDinterwordspacing

\bibitem{noauthor_depth_nodate}
\BIBentryALTinterwordspacing
``\BIBforeignlanguage{en-US}{Depth {Camera} {D405}}.'' [Online]. Available: \url{https://www.intelrealsense.com/depth-camera-d405/}
\BIBentrySTDinterwordspacing

\bibitem{todorov2012mujoco}
E.~Todorov, T.~Erez, and Y.~Tassa, ``Mujoco: A physics engine for model-based control,'' in \emph{2012 IEEE/RSJ International Conference on Intelligent Robots and Systems}.\hskip 1em plus 0.5em minus 0.4em\relax IEEE, 2012, pp. 5026--5033.

\bibitem{menagerie2022github}
\BIBentryALTinterwordspacing
K.~Zakka, Y.~Tassa, and {MuJoCo Menagerie Contributors}, ``{MuJoCo Menagerie: A collection of high-quality simulation models for MuJoCo},'' 2022. [Online]. Available: \url{http://github.com/google-deepmind/mujoco_menagerie}
\BIBentrySTDinterwordspacing

\bibitem{noauthor_aiortcaiortc_2024}
\BIBentryALTinterwordspacing
``aiortc/aiortc,'' Sept. 2024, original-date: 2018-02-23T22:05:16Z. [Online]. Available: \url{https://github.com/aiortc/aiortc}
\BIBentrySTDinterwordspacing

\bibitem{buss2004introduction}
S.~R. Buss, ``Introduction to inverse kinematics with jacobian transpose, pseudoinverse and damped least squares methods,'' \emph{IEEE Journal of Robotics and Automation}, vol.~17, no. 1-19, p.~16, 2004.

\bibitem{cadene2024lerobot}
R.~Cadene, S.~Alibert, A.~Soare, Q.~Gallouedec, A.~Zouitine, and T.~Wolf, ``Lerobot: State-of-the-art machine learning for real-world robotics in pytorch,'' \url{https://github.com/huggingface/lerobot}, 2024.

\bibitem{he2016deep}
K.~He, X.~Zhang, S.~Ren, and J.~Sun, ``Deep residual learning for image recognition,'' in \emph{Proceedings of the IEEE conference on computer vision and pattern recognition}, 2016, pp. 770--778.

\bibitem{mandlekar2021matters}
A.~Mandlekar, D.~Xu, J.~Wong, S.~Nasiriany, C.~Wang, R.~Kulkarni, L.~Fei-Fei, S.~Savarese, Y.~Zhu, and R.~Mart{\'\i}n-Mart{\'\i}n, ``What matters in learning from offline human demonstrations for robot manipulation,'' \emph{arXiv preprint arXiv:2108.03298}, 2021.

\end{thebibliography}

\end{document}